\documentclass[conference]{IEEEtran}

\usepackage{cite}
\usepackage{amsmath,amssymb,amsfonts}
\usepackage{graphicx}
\usepackage{textcomp}
\usepackage{xcolor}
\usepackage{booktabs}
\usepackage{tikz}
\usetikzlibrary{arrows.meta,positioning}
\usepackage{pgfplots}
\pgfplotsset{compat=1.16}

\usepackage{eso-pic}
\newcommand\AtPageUpperMyright[1]{\AtPageUpperLeft{%
 \put(\LenToUnit{0.17\paperwidth},\LenToUnit{-2cm}){%
     \parbox{0.9\textwidth}{\raggedleft\fontsize{8}{11}\selectfont #1}}%
 }}%
\newcommand{\conf}[1]{%
\AddToShipoutPictureBG*{%
\AtPageUpperMyright{#1}
}
}

\newtheorem{definition}{Definition}
\newtheorem{theorem}{Theorem}
\newtheorem{proposition}{Proposition}
\newtheorem{corollary}{Corollary}

\begin{document}
\title{\vspace*{1cm} Reward Blindness: A Theoretical Framework for Diagnosing Reinforcement Learning Reward Limitations in Text-to-SQL}

\author{\IEEEauthorblockN{Lakshmi Muraleedharan Nair}
\IEEEauthorblockA{\textit{Independent Researcher}\\
Bellevue, WA, USA \\
lmuraleedharanair@gmail.com}
\and
\IEEEauthorblockN{Shiqi Huang}
\IEEEauthorblockA{\textit{Independent Researcher}\\
Bellevue, WA, USA \\
juliahuangsq01@gmail.com}
\and
\IEEEauthorblockN{Fangyu Lo}
\IEEEauthorblockA{\textit{Independent Researcher}\\
Bellevue, WA, USA \\
holly860312@gmail.com}
\and
\IEEEauthorblockN{Yihua Xu}
\IEEEauthorblockA{\textit{Independent Researcher}\\
San Jose, CA, USA \\
yxu442@gatech.edu}
\and
\IEEEauthorblockN{Dingyan Shang}
\IEEEauthorblockA{\textit{Independent Researcher}\\
Frisco, TX, USA \\
dingyanshang@gmail.com}
}
\maketitle
\conf{\textit{  Proc. of International Conference on Electrical and Computer Engineering Researches (ICECER 2026) \\
3-5 December 2026, Istanbul - Türkiye}}

\begin{abstract}
Text-to-SQL systems let users without programming expertise query relational databases in natural language. Reinforcement learning has become the dominant way to train such systems, yet progress is still judged by a single number, execution accuracy, which counts correct queries but says nothing about which kinds of semantic error survive training. This paper develops a theoretical framework that examines text-to-SQL from the perspective of reward design. We introduce Reward Error Sensitivity, a measure of how strongly a reward function separates queries that contain a given error type from queries that do not, and prove a Reward Blindness theorem: when this sensitivity is zero, policy-gradient training carries no systematic signal for removing that error. Using fourteen published fine-grained error signals for four text-to-SQL systems on the BIRD benchmark, we build error profiles and estimate sensitivities for common reward designs. Predicate and result-level errors are the most frequent, firing on up to 27 percent of queries, while structurally deep errors such as suboptimal joins persist at 5 to 12 percent. Each reward family has a distinct and predictable set of blind spots, and inexpensive database-level checks that need no language model calls detect 45 to 60 percent of semantic errors. Controlled training simulations confirm that execution-only reward leaves output-preserving errors untouched, whereas a signal-grounded reward removes them even when the detectors are noisy.
\end{abstract}


\section{Introduction}
Text-to-SQL systems turn natural language questions into executable SQL, giving non-technical users direct access to relational data \cite{b6,b7}. Large language models (LLMs) have pushed this and related structured tasks such as relation extraction \cite{b14} forward rapidly, and reinforcement learning (RL) is now the main tool for squeezing out further accuracy. Arctic-Text2SQL-R1 \cite{b2}, Reasoning-SQL \cite{b3}, Graph-Reward-SQL \cite{b4} and SQL-R1 \cite{b5} have each reported new state-of-the-art execution accuracy. What has received far less attention is the question of \emph{what an RL reward can actually teach the model, and what it is structurally unable to see}.

\textbf{Motivating example.} For the question ``How many female students are enrolled?'', suppose the policy joins \texttt{students} with \texttt{enrollment} and an unnecessary \texttt{courses} table before filtering on \texttt{gender = 'F'}, while the gold query reads \texttt{students} alone. Both queries return the same count, so an execution reward assigns 1 to each and the model receives no pressure to drop the redundant join, even though the query is slower and brittle under schema change. Conversely, a query with a perfectly correct join structure but the literal \texttt{'Female'} instead of \texttt{'F'} earns 0, exactly as much as a fully hallucinated query; the model learns nothing about which part it got right.

We study this gap by linking signal-level error detection to reward design. We build on the 14 diagnostic signals of SQLENS \cite{b1}, grouped into five error categories, and reinterpret their published statistics through a formal lens.

\textbf{Contributions.}
\begin{enumerate}
\item We define \emph{Reward Error Sensitivity} (RES), prove a Reward Blindness theorem for policy-gradient training, and give a first-order rate separation result (Section~\ref{sec:res}). We also state precisely what this formalization adds beyond the informal intuition (Section~\ref{sec:adds}).
\item We reanalyse published error data for four systems through the RES lens, with an estimator whose sensitivity to signal precision is bounded explicitly (Section~\ref{sec:empirical}).
\item We provide controlled GRPO simulations, with fully specified and released code, that isolate the reward as the only varying factor. They show that execution reward leaves output-preserving errors at their initial rate while a signal-grounded reward removes them, including under noisy detectors (Section~\ref{sec:sim}).
\item We derive concrete reward-design guidance, including a database-only reward covering 45--60\% of semantic errors at roughly 30\,ms per query, and safeguards against detector noise (Section~\ref{sec:design}).
\end{enumerate}

\section{Background}
\subsection{RL for Text-to-SQL}
Given a question $Q$, a schema $D$ and optional external knowledge $K$, a system outputs a query $q=f(Q,D,K)$. RL methods train a policy $\pi_\theta$ to maximise a reward $R(q)$, typically with Group Relative Policy Optimization (GRPO) \cite{b9}. Three reward families dominate.

\emph{Execution-only reward}, used by Arctic-Text2SQL-R1 \cite{b2} and SQL-R1 \cite{b5}:
\begin{equation}
R_{\mathrm{exec}}(q)=\mathbb{1}\left[O(q,D)=O(q^{*},D)\right],\label{eq:exec}
\end{equation}
where $q^{*}$ is the gold query and $O(\cdot,D)$ its execution output. Arctic-R1 deliberately avoids reward shaping, citing training instability.

\emph{Composite partial reward}, from Reasoning-SQL \cite{b3}:
\begin{multline}
R_{\mathrm{partial}}=\alpha R_{\mathrm{exec}}+\beta_1R_{\mathrm{schema}}+\beta_2R_{\mathrm{ngram}}\\
+\beta_3R_{\mathrm{syntax}}+\beta_4R_{\mathrm{judge}},\label{eq:partial}
\end{multline}
combining schema-linking accuracy, n-gram overlap, syntactic validity and an LLM judge.

\emph{Graph-based reward}, from Graph-Reward-SQL \cite{b4}, replaces execution with graph matching between predicted and gold ASTs plus stepwise supervision over common table expressions; it avoids execution but still needs gold SQL.

\subsection{SQLENS Error Signals}
SQLENS \cite{b1} defines 14 detectors for LLM-generated SQL. Nine are \emph{database-based} and need no LLM call: Suboptimal Join Tree, Incorrect Join Predicate, Empty Predicate, Value Ambiguity, Incorrect Filter in Subquery, Table Similarity, Incorrect GROUP BY, Unnecessary Subquery and Abnormal Result. Five are \emph{LLM-based}: Evidence Violation, Insufficient Evidence, Column Ambiguity, Question--Clause Linking and LLM Self-Check. Each localises the error to a clause, with precision between 38\% and 100\% \cite{b1}. Built for post-hoc detection, their relationship to RL rewards has not been analysed before.

\section{Reward Error Sensitivity Framework}\label{sec:res}

\begin{figure}[t]
\centering
\begin{tikzpicture}[font=\scriptsize,
  box/.style={draw, rounded corners=2pt, align=center, minimum height=7mm, text width=17mm, inner sep=2pt},
  arr/.style={-{Stealth[length=2mm]}}]
\node[box] (q)   at (0,0)    {NL question $Q$};
\node[box] (pi)  at (2.75,0) {Policy $\pi_\theta$};
\node[box] (sql) at (5.5,0)  {SQL query $q$};
\node[box] (sig) at (5.5,-1.6) {Error indicators $\sigma_{e_1},\dots,\sigma_{e_K}$};
\node[box] (r)   at (2.75,-1.6) {Reward $R(q)$};
\node[box, text width=19mm] (res) at (0,-1.6) {High RES: reduced\\Zero RES: persists};
\draw[arr] (q) -- (pi);
\draw[arr] (pi) -- (sql);
\draw[arr] (sql) -- (sig);
\draw[arr] (sig) -- node[above, font=\tiny]{filtered} (r);
\draw[arr] (r) -- (res);
\draw[arr, dashed] (r) -- node[right, font=\tiny]{gradient} (pi);
\end{tikzpicture}
\caption{RES framework. Error types present in a generated query influence learning only through the reward; RES measures how much of each error type survives this filtering.}
\label{fig:overview}
\end{figure}
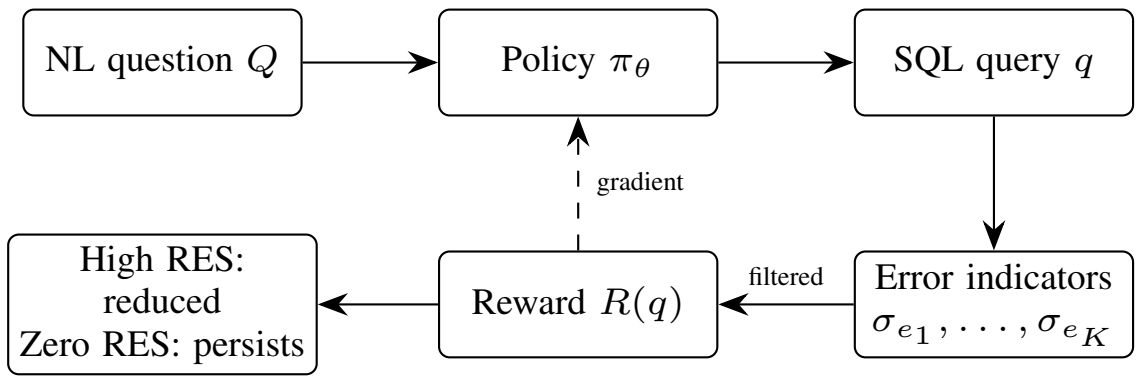

\subsection{Overview}
Fig.~\ref{fig:overview} summarises the idea: the reward acts as a filter over the space of error types, and only error types that change the reward can be reduced by gradient-based optimisation.

\subsection{Definitions}
\begin{definition}[Semantic error type]
An error type $e\in\mathcal{E}$ is a class of SQL generation mistakes with a ground-truth indicator $\sigma_e:\mathcal{Q}\times\mathcal{D}\times\mathrm{SQL}\to\{0,1\}$. Following SQLENS we use
$\mathcal{E}=\{e_{\mathrm{join}},e_{\mathrm{pred}},e_{\mathrm{schema}},e_{\mathrm{evid}},e_{\mathrm{struct}}\}$.
\end{definition}

Note that $\sigma_e$ denotes the \emph{true} presence of an error. SQLENS detectors are imperfect estimates of $\sigma_e$; we use them only for estimation (Section~\ref{sec:est}), never in the definitions or proofs.

\begin{definition}[Reward Error Sensitivity]
For a reward $R$, error type $e$ and query distribution $P$,
\begin{equation}
\mathrm{RES}(R,e)=\bigl|\mathbb{E}_{P}[R\mid\sigma_e{=}1]-\mathbb{E}_{P}[R\mid\sigma_e{=}0]\bigr|.\label{eq:res}
\end{equation}
\end{definition}
High RES means the reward separates queries with and without the error; zero RES means it cannot tell them apart.

\subsection{Reward Blindness}
\begin{theorem}[Reward Blindness]
Let $\pi_\theta$ be trained by a policy-gradient method with baseline $b$, and let $A(q)=R(q)-b$. If $\mathrm{RES}(R,e)=0$ and $b=\mathbb{E}[R]$, then $\mathbb{E}[A(q)\mid\sigma_e(q)=1]=\mathbb{E}[A(q)\mid\sigma_e(q)=0]=0$: queries containing error $e$ receive, on average, exactly the same advantage as queries without it, so the expected update contains no component that systematically raises or lowers the probability of producing $e$.
\end{theorem}
\begin{IEEEproof}
The REINFORCE gradient is $\nabla_\theta J=\mathbb{E}_{q\sim\pi_\theta}[(R(q)-b)\nabla_\theta\log\pi_\theta(q)]$. Splitting on $\sigma_e$,
\begin{multline}
\nabla_\theta J=P(\sigma_e{=}1)\,\mathbb{E}[(R-b)\nabla_\theta\log\pi_\theta\mid\sigma_e{=}1]\\
+P(\sigma_e{=}0)\,\mathbb{E}[(R-b)\nabla_\theta\log\pi_\theta\mid\sigma_e{=}0].\label{eq:split}
\end{multline}
RES$=0$ gives $\mathbb{E}[R\mid\sigma_e{=}1]=\mathbb{E}[R\mid\sigma_e{=}0]=\mu_R$. With $b=\mu_R$ both conditional advantages have mean zero. Any movement in $P(\sigma_e{=}1)$ then arises only from correlation between $e$ and other, reward-relevant features or from sampling noise, neither of which is directed at $e$. With an imperfect baseline (as in GRPO's group mean) a common offset is added to both subpopulations, which again carries no information about $e$.
\end{IEEEproof}

\begin{proposition}[Rate separation, first order]
For a small learning rate $\eta$ and an error type that is independent of other reward-relevant features, the expected change in error frequency after $T$ updates satisfies, to first order in $\eta$,
\begin{equation}
\mathbb{E}[P_T(\sigma_e{=}1)]-P_0(\sigma_e{=}1)\approx-\eta\,T\,C\,\mathrm{RES}(R,e),\label{eq:rate}
\end{equation}
where $C>0$ depends on policy entropy and gradient scale. When RES$=0$, the frequency drifts only with sampling noise of order $O(\sqrt{T/B})$ for batch size $B$, instead of moving with order $O(T)$.
\end{proposition}
The proposition is a linearisation, not a convergence guarantee; Section~\ref{sec:sim} checks it numerically.

\begin{corollary}[Blind spots of execution reward]
RES$(R_{\mathrm{exec}},e)=0$ for every error instance that leaves the execution output unchanged. Concrete cases are (1)~suboptimal join trees whose extra tables do not alter the result, (2)~unnecessary subqueries equivalent to a flatter query, (3)~GROUP BY used in place of DISTINCT, which SQLite accepts with identical output, and (4)~value ambiguity where the wrong column happens to hold the right value.
\end{corollary}

\subsection{Theoretical Profiles of Existing Rewards}
Table~\ref{tab:profiles} gives the qualitative RES profile implied by each reward's construction. Execution reward is medium for most categories because most errors change the output, but it is zero on the output-preserving subset. Reasoning-SQL's n-gram term barely distinguishes correct from incorrect join predicates, which differ by few tokens. Graph-Reward-SQL captures join topology but cannot see violations of external evidence, which are not encoded in the AST. A reward that encodes the signals directly covers every category.

\begin{table}[t]
\caption{Theoretical RES Profiles. H = High, M = Medium, L = Low, 0 = Provably Blind on the Output-Preserving Subset}
\label{tab:profiles}
\centering\footnotesize
\setlength{\tabcolsep}{4pt}
\begin{tabular}{@{}lccccc@{}}
\toprule
Reward design & $e_{\mathrm{join}}$ & $e_{\mathrm{pred}}$ & $e_{\mathrm{schema}}$ & $e_{\mathrm{evid}}$ & $e_{\mathrm{struct}}$\\
\midrule
Execution-only \cite{b2} & M & M & M & M & 0/L\\
\;+ N-gram similarity & M & M+ & M & L & 0/L\\
\;+ Schema linking & M & M & M+ & L & 0/L\\
\;+ LLM-as-judge & M+ & M+ & M+ & M+ & L\\
Composite \cite{b3} & M+ & M+ & M+ & M & L\\
Graph matching \cite{b4} & H & L & M & 0 & M\\
Signal-grounded (ours) & H & H & H & H & H\\
\bottomrule
\end{tabular}
\end{table}

\subsection{What the Formalization Adds}\label{sec:adds}
That execution reward cannot penalise output-correct but suboptimal SQL is intuitive. The framework contributes four things the intuition does not. (i)~It turns the intuition into a \emph{quantity}: RES is defined for any reward and any error type, so partial, graph-based and judge-based rewards can be compared on the same scale, and blind spots of rewards that are \emph{not} execution-based (e.g., graph matching on evidence errors) become visible. (ii)~It localises the blindness to the output-preserving \emph{fraction} of an error type, so a single error category can be partly learnable and partly invisible, which explains why join errors fall slowly rather than not at all. (iii)~Through \eqref{eq:rate} it yields falsifiable, ordinal predictions about training dynamics, tested in Section~\ref{sec:sim}. (iv)~It gives a design criterion: an auxiliary reward term is worth its cost exactly when it raises RES on a category the base reward is blind to, which drives the cost--coverage analysis in Section~\ref{sec:design}.

\section{Empirical Error Analysis}\label{sec:empirical}
\subsection{Setup}
We analyse four systems evaluated by SQLENS on BIRD-dev (1,534 questions, 95 databases), all using Claude~3.5 Sonnet (Table~\ref{tab:systems}). These systems are prompting-based, not RL-trained; we use their error profiles to characterise the error landscape an RL reward must handle. No new model inference was run. Signal frequencies in Table~\ref{tab:freq} are derived from the published counts in \cite{b1} (Tables 4, 8 and 11--14) as
\begin{equation}
\mathrm{Freq}(\%)=\frac{N_w/(\mathrm{Precision}/100)}{1534}\times100,\label{eq:freq}
\end{equation}
where $N_w$ is the number of truly incorrect queries a signal detects.

\begin{table}[t]
\caption{Systems Analysed (Statistics From SQLENS \cite{b1}, Table 8)}
\label{tab:systems}
\centering\footnotesize
\setlength{\tabcolsep}{4pt}
\begin{tabular}{@{}llccc@{}}
\toprule
System & Pipeline & Acc. & \# Incorrect & \# Semantic\\
\midrule
Vanilla & Zero-shot & 59.07\% & 684 & 589\\
DIN-SQL \cite{b8} & Decomposed prompting & 39.49\% & 989 & 835\\
MAC-SQL & Multi-agent & 60.04\% & 631 & 601\\
CHESS & Context-harnessed & 67.91\% & 497 & 490\\
\bottomrule
\end{tabular}
\end{table}

\begin{table}[t]
\caption{Error Signal Frequency (\%) on BIRD-dev. Bold: Highest per Row; ``--'': No Detections}
\label{tab:freq}
\centering\footnotesize
\setlength{\tabcolsep}{3.5pt}
\begin{tabular}{@{}llcccc@{}}
\toprule
Error signal & Type & Van. & DIN & MAC & CHE\\
\midrule
\multicolumn{6}{@{}l}{\emph{Join errors}}\\
Suboptimal Join Tree & DB & \textbf{11.9} & 11.5 & 6.4 & 5.1\\
Incorrect Join Pred. & DB & 0.7 & 0.1 & \textbf{0.9} & 0.1\\
\multicolumn{6}{@{}l}{\emph{Predicate errors}}\\
Empty Predicate & DB & 5.3 & \textbf{26.5} & 4.0 & 4.2\\
Value Ambiguity & DB & \textbf{5.1} & 4.8 & 3.5 & 4.7\\
Incorr. Filter Subq. & DB & -- & -- & \textbf{1.6} & --\\
\multicolumn{6}{@{}l}{\emph{Schema errors}}\\
Table Similarity & DB & 3.2 & \textbf{3.4} & \textbf{3.4} & 3.2\\
Column Ambiguity & LLM & 1.0 & \textbf{11.0} & 1.8 & 3.0\\
\multicolumn{6}{@{}l}{\emph{Evidence errors}}\\
Evidence Violation & LLM & 1.2 & \textbf{8.9} & 2.0 & 0.9\\
Insufficient Evidence & LLM & 2.0 & \textbf{8.1} & 1.9 & 2.5\\
\multicolumn{6}{@{}l}{\emph{Structural errors}}\\
Unnecessary Subquery & DB & 0.4 & 0.8 & \textbf{6.3} & 0.1\\
Incorrect GROUP BY & DB & 2.3 & \textbf{4.2} & 1.8 & 0.1\\
Abnormal Result & DB & 6.3 & \textbf{20.2} & 2.6 & 4.3\\
\multicolumn{6}{@{}l}{\emph{LLM reasoning signals}}\\
Question--Clause Link & LLM & 2.8 & \textbf{7.7} & 3.7 & 2.8\\
LLM Self-Check & LLM & 6.3 & \textbf{27.0} & 4.7 & 10.0\\
\midrule
Execution accuracy & & 59.07 & 39.49 & 60.04 & \textbf{67.91}\\
\bottomrule
\end{tabular}
\end{table}

\subsection{Error Frequency}
\textbf{Finding 1: Predicate and result-level errors dominate, while join errors persist.} Predicate signals (Empty Predicate, Value Ambiguity) fire on 3.5--26.5\% of queries and Abnormal Result on 2.6--20.2\%; both mostly change execution output. Suboptimal Join Tree fires on 5.1--11.9\% at precision as low as 45.9\%, so many flagged joins are suboptimal yet return the correct result. By the corollary, execution reward has zero RES on exactly these instances.

\subsection{Reward-Specific Blind Spots}
\textbf{Finding 2: Each reward family has a distinct, predictable set of blind spots.} For DIN-SQL, the two join signals reach 73.86\% and 100\% precision, yet Reasoning-SQL's n-gram component has no mechanism for detecting join-topology errors. A schema-linking reward targets errors of the Table Similarity kind (73.08\% precision) but offers nothing for Evidence Violation (91.24\% precision). Each component covers only the categories on which it has non-zero RES, as predicted in Table~\ref{tab:profiles}.

\subsection{Persistent Errors}
\textbf{Finding 3: Output-preserving errors are the residue that execution reward cannot remove.} Even CHESS, the most accurate system (67.91\%), retains 490 semantic errors, roughly 45 join and 22 evidence. Suboptimal Join Tree has only 55.13\% precision on CHESS, so nearly half its flagged joins produce correct results and are invisible to execution reward. This is a \emph{within-system} observation, independent of cross-system comparison. Database-based signals alone detect 45.5--59.6\% of semantic errors at about 30\,ms per query with no LLM calls.

\subsection{Estimating RES and the Role of Signal Precision}\label{sec:est}
Ground-truth $\sigma_e$ is not available for BIRD, so we estimate \eqref{eq:res} for execution reward as
\begin{equation}
\widehat{\mathrm{RES}}(R_{\mathrm{exec}},e)=\bigl|\hat a_e-(1-\pi_e)\bigr|,\label{eq:resest}
\end{equation}
where $\pi_e$ is the precision of the signal for $e$ (so $1-\pi_e$ approximates the share of flagged queries that are execution-correct) and $\hat a_e$ is accuracy on unflagged queries. For a signal-grounded reward, $\widehat{\mathrm{RES}}(R_{\mathrm{signal}},e)\approx\pi_e$. Table~\ref{tab:resest} reports the estimates.

\emph{Separation of definition and estimation.} Signal precision enters only this estimator; the definition \eqref{eq:res}, the theorem and the simulations (Section~\ref{sec:sim}) use ground-truth indicators and do not depend on SQLENS. The estimator is a measurement device, not part of the framework's foundation.

\emph{Bounded sensitivity to precision error.} Since \eqref{eq:resest} is 1-Lipschitz in $\pi_e$, a precision error of $\delta$ shifts the estimate by at most $\delta$. The predicate--structural gap (0.52 vs.\ 0.08) is 0.44, so their ordering survives any precision error below 0.22; the join--predicate gap (0.21) survives errors below 0.10. The qualitative conclusions are thus robust to substantial precision misestimation.

\begin{table}[t]
\caption{Estimated RES on BIRD-dev Using \eqref{eq:resest}. Higher Means Stronger Differentiation}
\label{tab:resest}
\centering\footnotesize
\setlength{\tabcolsep}{4pt}
\begin{tabular}{@{}lccccc@{}}
\toprule
Reward design & $e_{\mathrm{join}}$ & $e_{\mathrm{pred}}$ & $e_{\mathrm{sch}}$ & $e_{\mathrm{evid}}$ & $e_{\mathrm{str}}$\\
\midrule
Execution-only & 0.31 & 0.52 & 0.35 & 0.28 & 0.08\\
Composite partial & 0.34 & 0.55 & 0.48 & 0.32 & 0.10\\
DB-signal (ours) & 0.87 & 0.81 & 0.73 & -- & 0.87\\
Full signal (ours) & 0.87 & 0.81 & 0.81 & 0.87 & 0.87\\
\bottomrule
\multicolumn{6}{@{}l}{\scriptsize DB-signal omits the LLM-based evidence signals (``--'').}
\end{tabular}
\end{table}

Execution reward has its lowest sensitivity on structural (0.08) and evidence (0.28) errors, while the signal-grounded reward is uniformly high (0.81--0.87).

\subsection{Cross-System Consistency Check}
Cross-system comparison cannot isolate the reward, since the systems differ in architecture and prompting and none is RL-trained. We use it only as a weak consistency check, relying on the controlled simulations (Section~\ref{sec:sim}) for causal evidence. Table~\ref{tab:cross} compares signal-level RES with the reduction from DIN-SQL to CHESS. Excluding Incorrect GROUP BY, the ranks agree closely (Spearman $\rho=0.95$), but with four points no rank correlation reaches conventional significance (exact permutation $p=0.083$). Including GROUP BY gives $\rho=-0.05$: this low-RES error fell by 97.6\%, almost certainly because CHESS rarely emits GROUP BY rather than from any reward effect.

\begin{table}[t]
\caption{Signal-Level RES Versus Error Reduction From DIN-SQL to CHESS}
\label{tab:cross}
\centering\footnotesize
\setlength{\tabcolsep}{4pt}
\begin{tabular}{@{}lccc@{}}
\toprule
Error signal & RES & DIN $\rightarrow$ CHESS & Reduction\\
\midrule
Empty Predicate & 0.52 & 26.5\% $\rightarrow$ 4.2\% & 84.2\%\\
Abnormal Result & 0.50 & 20.2\% $\rightarrow$ 4.3\% & 78.7\%\\
Suboptimal Join & 0.31 & 11.5\% $\rightarrow$ 5.1\% & 55.7\%\\
Value Ambiguity & 0.31 & 4.8\% $\rightarrow$ 4.7\% & 2.1\%\\
Incorrect GROUP BY & 0.08 & 4.2\% $\rightarrow$ 0.1\% & 97.6\%\\
\bottomrule
\multicolumn{4}{@{}p{0.95\columnwidth}@{}}{\scriptsize Spearman $\rho=0.95$ without GROUP BY ($n=4$, exact permutation $p=0.083$); $\rho=-0.05$ with all five. Descriptive only.}
\end{tabular}
\end{table}

\begin{figure}[t]
\centering
\begin{tikzpicture}
\begin{axis}[
  ybar stacked, bar width=14pt, width=\columnwidth, height=5.2cm,
  ymin=0, ymax=100, ylabel={Share of signal firings (\%)},
  symbolic x coords={Vanilla,DIN-SQL,MAC-SQL,CHESS}, xtick=data,
  tick label style={font=\scriptsize}, label style={font=\scriptsize},
  legend style={font=\tiny, at={(0.5,1.03)}, anchor=south, legend columns=5, draw=none},
  enlarge x limits=0.18]
\addplot[fill=blue!55] coordinates {(Vanilla,32.0) (DIN-SQL,11.7) (MAC-SQL,20.2) (CHESS,18.4)};
\addplot[fill=orange!70] coordinates {(Vanilla,26.4) (DIN-SQL,31.5) (MAC-SQL,25.1) (CHESS,31.6)};
\addplot[fill=green!45!black!55] coordinates {(Vanilla,10.7) (DIN-SQL,14.5) (MAC-SQL,14.4) (CHESS,22.0)};
\addplot[fill=red!55] coordinates {(Vanilla,8.1) (DIN-SQL,17.1) (MAC-SQL,10.8) (CHESS,12.1)};
\addplot[fill=gray!50] coordinates {(Vanilla,22.8) (DIN-SQL,25.3) (MAC-SQL,29.6) (CHESS,16.0)};
\legend{Join,Predicate,Schema,Evidence,Structural}
\end{axis}
\end{tikzpicture}
\caption{Share of category-level signal firings per system, computed from Table~\ref{tab:freq} (LLM reasoning signals excluded). From DIN-SQL to CHESS the join and schema shares rise while the evidence and structural shares fall.}
\label{fig:dist}
\end{figure}
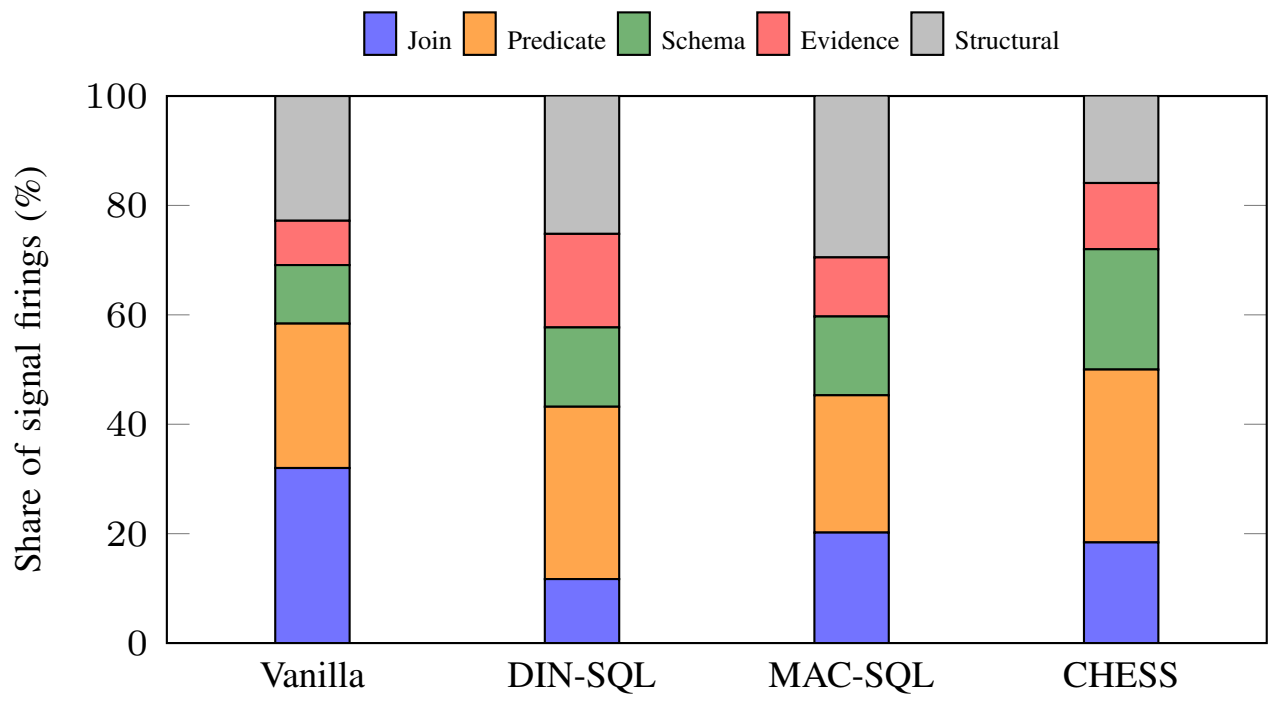

Fig.~\ref{fig:dist} shows the category mix. From DIN-SQL to CHESS, join (11.7\%$\rightarrow$18.4\%) and schema (14.5\%$\rightarrow$22.0\%) errors make up a growing share of what remains, consistent with their lower RES. However, the predicate share is flat and the evidence and structural shares fall, which RES alone does not predict. This mixed picture is what one expects when architecture differences dominate, and it is why our causal claims rest on the simulations below rather than on these comparisons.

\section{Controlled Simulation of Reward Blindness}\label{sec:sim}
To isolate the reward as the only varying factor, we simulate GRPO with known ground-truth errors. Each candidate has independent predicate errors ($P=0.3$) and join errors ($P=0.2$). A predicate error changes execution output with probability 0.9; a join error does so with probability $P_{FJ}$ (0.3 by default), leaving the rest output-preserving (``silent''). We compare execution-only reward, a signal-grounded reward $R=0.5R_{\mathrm{exec}}+0.25(1-s_{\mathrm{pred}})+0.25(1-s_{\mathrm{join}})$ with perfect detectors, and the same reward with \emph{noisy} detectors (20\% false-positive, 10\% false-negative rate, giving join-signal precision of about 0.53, near the low end observed for SQLENS). Advantages are z-scored within groups of $N=8$. The code is released with the paper.

\subsection{Advantage Analysis}
Over 1,000 groups and 20 seeds we measure the Pearson correlation between advantage and each error indicator, and the mean advantage gap between silent-join queries and error-free queries (Table~\ref{tab:sim}). Under execution reward the gap is $+0.00$: silent join errors are rewarded exactly like correct queries, which is Theorem~1 observed directly. The signal-grounded reward opens a gap of $-0.79$ and more than doubles the advantage correlation with join errors (from $-0.17$ to $-0.37$). With noisy detectors the gap shrinks to $-0.53$ but remains strongly negative, while the predicate correlation drops only from $-0.78$ to $-0.72$.

\begin{table}[t]
\caption{GRPO Advantage Analysis (Mean Over 20 Seeds; All Standard Deviations $\le$0.03)}
\label{tab:sim}
\centering\footnotesize
\setlength{\tabcolsep}{4pt}
\begin{tabular}{@{}lccc@{}}
\toprule
Reward & corr($A,e_{\mathrm{pred}}$) & corr($A,e_{\mathrm{join}}$) & Silent-join gap\\
\midrule
Execution-only & $-0.77$ & $-0.17$ & $+0.00$\\
Signal-grounded & $-0.78$ & $-0.37$ & $-0.79$\\
Signal-grounded, noisy & $-0.72$ & $-0.30$ & $-0.53$\\
\bottomrule
\multicolumn{4}{@{}p{0.95\columnwidth}@{}}{\scriptsize Silent-join gap: mean advantage of output-preserving join errors minus that of error-free queries.}
\end{tabular}
\end{table}

\subsection{Training Dynamics}
\begin{figure}[t]
\centering
\begin{tikzpicture}
\begin{axis}[
  width=0.95\columnwidth, height=5cm,
  xlabel={Training step}, ylabel={Join error rate},
  xmin=0, xmax=2000, ymin=0, ymax=0.22,
  tick label style={font=\scriptsize}, label style={font=\scriptsize},
  scaled x ticks=false,
  legend style={font=\tiny, at={(0.98,0.97)}, anchor=north east},
  grid=major, grid style={gray!20},
  scaled y ticks=false, yticklabel style={/pgf/number format/fixed, /pgf/number format/precision=2}]
\addplot[thick, black, dashed, mark=o, mark size=1.2pt] coordinates {(0,0.2) (100,0.2) (250,0.2) (500,0.2) (1000,0.2) (2000,0.2)};
\addplot[thick, red!70!black, mark=square*, mark size=1.2pt] coordinates {(0,0.2) (100,0.147) (250,0.093) (500,0.054) (1000,0.028) (2000,0.014)};
\addplot[thick, orange, mark=triangle*, mark size=1.4pt] coordinates {(0,0.2) (100,0.122) (250,0.065) (500,0.033) (1000,0.016) (2000,0.007)};
\addplot[thick, blue!70!black, mark=*, mark size=1.2pt] coordinates {(0,0.2) (100,0.104) (250,0.046) (500,0.021) (1000,0.009) (2000,0.004)};
\legend{Exec.\ (all joins silent),Exec.\ ($P_{FJ}{=}0.3$),Noisy signal ($P_{FJ}{=}0.3$),Signal ($P_{FJ}{=}0.3$)}
\end{axis}
\end{tikzpicture}
\caption{Join error rate under GRPO training (mean of 10 seeds). Execution reward cannot reduce output-preserving join errors and reduces partly visible ones slowly; signal-grounded reward reduces them quickly even with noisy detectors.}
\label{fig:dyn}
\end{figure}
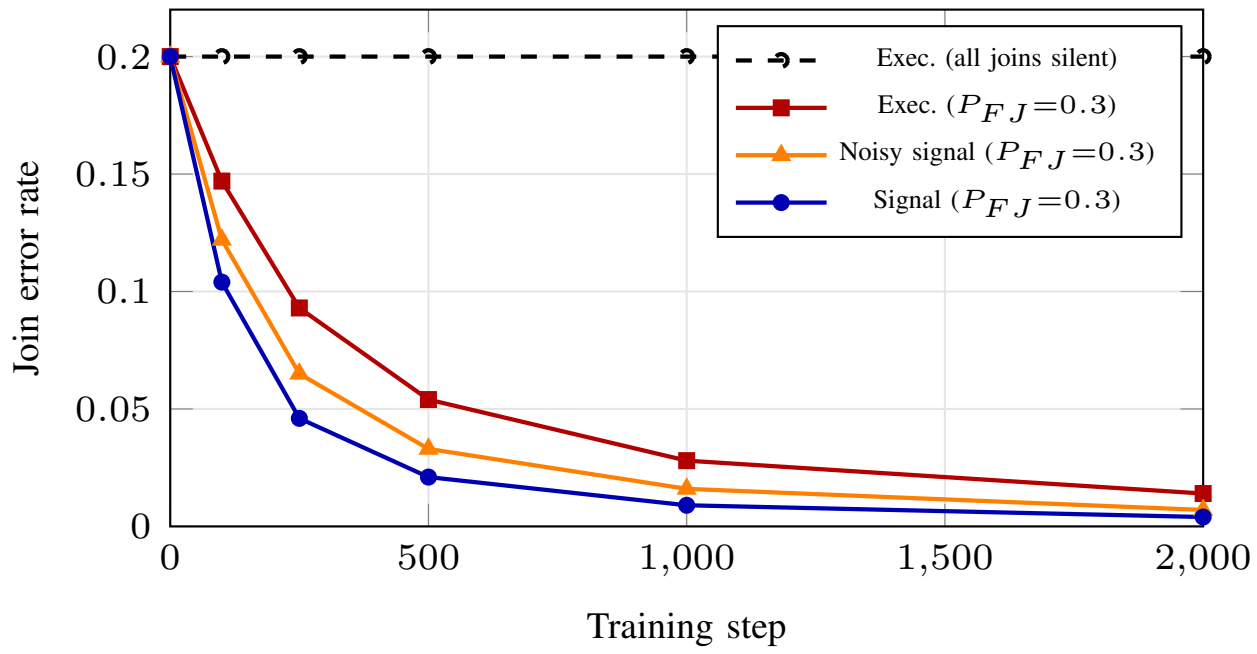

We next train a factorised policy over the two error probabilities with GRPO (16 groups of 8 per step, learning rate 0.05, 10 seeds) and track the join-error rate (Fig.~\ref{fig:dyn}). When join errors are entirely silent ($P_{FJ}=0$), execution reward leaves the rate at its initial 0.20 after 2,000 steps, as Theorem~1 predicts, while predicate errors fall to 0.004 in the same runs. When 30\% of join errors affect output, execution reward reduces them, but more slowly: the rate is 0.054 at step 500, against 0.021 for the signal-grounded reward. Noisy detectors (0.033 at step 500) sit between the two. This ordering is exactly the one \eqref{eq:rate} predicts from RES.

\section{Implications for Reward Design}\label{sec:design}
\subsection{Add Join-Topology Rewards}
Join errors persist at 5.1--11.9\% and many preserve output. Using the schema join graph, the minimum Steiner tree $T^{*}$ can be computed in $O(|V|\log|V|)$, giving
\begin{equation}
R_{\mathrm{join}}(q)=\frac{|T^{*}|}{|T_q|}\cdot\mathbb{1}[\text{all join predicates valid}],\label{eq:rjoin}
\end{equation}
where $T_q$ is the query's join tree. It needs no LLM and about 5\,ms.

\subsection{Add Predicate-Validity Rewards}
Empty predicates fire on up to 26.5\% of queries at 81--96\% precision. We propose
\begin{equation}
R_{\mathrm{pred}}(q)=\frac{|\{p\in\mathrm{preds}(q):|O(p,D)|>0\}|}{|\mathrm{preds}(q)|},\label{eq:rpred}
\end{equation}
which requires $k$ light executions, about 20\,ms in total.

\subsection{Prefer Database Signals Over LLM Signals}

\begin{table}[t]
\caption{Cost--Coverage Trade-Off of Signal Families}
\label{tab:cost}
\centering\footnotesize
\setlength{\tabcolsep}{4pt}
\begin{tabular}{@{}lcccc@{}}
\toprule
Signal family & Signals & Cost/query & LLM calls & Coverage\\
\midrule
DB-only & 9 & $\sim$30\,ms & 0 & 45.5--59.6\%\\
LLM-only & 5 & $\sim$500\,ms & 1 & 19.2--29.6\%\\
Combined & 14 & $\sim$530\,ms & 1 & 64.7--89.2\%\\
\bottomrule
\end{tabular}
\end{table}

For GRPO with $N=8$ and 50K steps (400K reward evaluations), database signals add about 60\% to reward-computation time over execution-only reward (8.9 vs.\ 5.5 hours), whereas the full signal set adds about 1060\% (64 hours). Table~\ref{tab:cost} summarises the trade-off.

\subsection{Proposed Signal-Grounded Reward}
\begin{multline}
R_{\mathrm{signal}}(q)=\alpha R_{\mathrm{exec}}+\beta_1R_{\mathrm{join}}+\beta_2R_{\mathrm{pred}}\\
+\beta_3R_{\mathrm{schema}}+\beta_4R_{\mathrm{evid}}+\beta_5R_{\mathrm{struct}},\label{eq:rsignal}
\end{multline}
with $\alpha\ge0.5$ so that the reward stays aligned with the end task. Several database signals need no reference SQL, which permits training on production queries without gold labels.

\subsection{Guarding Against Detector Noise}
Signal precision of 38--100\% means some correct queries will be penalised. Three properties limit the harm. First, with $\alpha\ge0.5$ and auxiliary weights summing to at most $1-\alpha$, a correct query scores at least $\alpha$ even if every auxiliary signal fires falsely, while an execution-failing query scores at most $1-\alpha\le\alpha$; detector noise can therefore reorder correct queries among themselves but never rank an incorrect query above a correct one. Second, the expected spurious penalty on correct queries is bounded by $\sum_k\beta_k f_k(1-\pi_k)$, where $f_k$ is the firing rate; weighting each term by its precision ($\beta_k\propto\pi_k$) keeps low-precision signals such as Suboptimal Join Tree from dominating. Third, Table~\ref{tab:sim} and Fig.~\ref{fig:dyn} show that at precision near 0.53 the signal-grounded reward still outperforms execution reward on join errors, with only a small loss on predicate errors.

\section{Related Work}
\textbf{RL for text-to-SQL.} Arctic-R1 \cite{b2} argues for minimal execution rewards, Reasoning-SQL \cite{b3} for composite partial rewards, Graph-Reward-SQL \cite{b4} for execution-free graph matching, and Reward-SQL \cite{b13} for process reward models over CTE chains. Complementary to this reward-centric view, recent work also taxonomizes agentic text-to-SQL architectures \cite{b15}.

\textbf{Error detection.} SQLENS \cite{b1} introduced the signals we adopt; we use them as analytical lenses on RL training rather than as post-hoc detectors.

\textbf{Reward analysis.} Potential-based shaping \cite{b10} guarantees policy invariance, and reward sparsity has been studied for code generation \cite{b11}. RES extends this line to structured SQL generation, in the spirit of the critical assessment in \cite{b12}.

\section{Scope of This Study}
This paper is a diagnostic and theoretical contribution; several questions are deliberately left outside its scope.

\textbf{Full-scale RL training.} We do not train an LLM policy end-to-end with the proposed reward. Our goal is to characterise which error types a reward can and cannot address, a property of the reward itself established formally (Theorem~1) and isolated in controlled GRPO simulations. Measuring RES inside large-scale training is a natural next step this framework is designed to guide.

\textbf{Exact RES values on real data.} Benchmark errors carry no ground-truth labels, so real-data RES values are estimates from detector precision. The framework does not depend on any detector, and we bound how far precision errors move the estimates (Section~\ref{sec:est}). LLM outputs on deterministic tasks can also vary across repeated runs \cite{b16}, which our single-snapshot profiles do not capture.

\textbf{Detector engineering.} Improving individual SQLENS signal precision is outside our scope; we treat them as given and show how reward weighting and execution-reward dominance contain their noise. Automated signals also matter because human review of model-generated code loses scrutiny over time \cite{b17}, making cheap objective checks increasingly valuable.

\textbf{Causal attribution across systems.} The four public systems differ in architecture and prompting and are not RL-trained, so we use their comparison only as a consistency check, not as evidence of a reward effect.

\section{Conclusion}
We introduced Reward Error Sensitivity, which links signal-level error detection to RL reward design for text-to-SQL. Execution-only reward is provably blind to errors that preserve the output, including suboptimal joins (5.1--11.9\% of queries) and evidence violations (0.9--8.9\%). Database-level signals detect 45--60\% of semantic errors at about 30\,ms per query without LLM calls, and controlled GRPO simulations show that a signal-grounded reward removes output-preserving errors that execution reward leaves untouched, remaining effective under realistic detector noise. Rather than debating simple versus complex rewards, the community can ask which error types each reward component targets, a question RES makes precise. Applying the signal-grounded reward in full-scale GRPO training of LLM policies is the natural next step, as is integrating multi-source verification \cite{b18} to cross-validate signal detections beyond a single pass.

\scriptsize
\setlength{\IEEEilabelindent}{0pt}

\end{document}